\documentclass[letterpaper]{article} 
\usepackage{aaai24}  
\usepackage{times}  
\usepackage{helvet}  
\usepackage{courier}  
\usepackage[hyphens]{url}  
\usepackage{graphicx} 
\usepackage{natbib}  
\usepackage{caption} 
\usepackage{algorithm}
\usepackage{algorithmic}

\usepackage{newfloat}
\usepackage{listings}
\DeclareCaptionStyle{ruled}{labelfont=normalfont,labelsep=colon,strut=off} 
\floatstyle{ruled}
\newfloat{listing}{tb}{lst}{}
\floatname{listing}{Listing}
\usepackage{xcolor}

\newcommand{\rom}[1]{\lowercase\expandafter{\romannumeral#1\relax}}

\usepackage{amsmath,amsfonts,bm}

\def\eqref#1{equation~\ref{#1}}

\def\1{\bm{1}}

\def\rvz{{\mathbf{z}}}

\DeclareMathAlphabet{\mathsfit}{\encodingdefault}{\sfdefault}{m}{sl}
\SetMathAlphabet{\mathsfit}{bold}{\encodingdefault}{\sfdefault}{bx}{n}

\DeclareMathOperator*{\argmin}{arg\,min}

\usepackage{multicol}
\usepackage{multirow}
\usepackage{enumerate}
\usepackage{algorithm}
\usepackage{algorithmic}
\usepackage{graphicx}
\usepackage{microtype}
\usepackage{graphicx}
\usepackage{caption}
\usepackage{subcaption}
\usepackage{booktabs} 
\usepackage{mathtools}
\usepackage{pifont}

\title{Feature Unlearning for Pre-trained GANs and VAEs}
\author {
    Saemi Moon\textsuperscript{\rm 1},
    Seunghyuk Cho\textsuperscript{\rm 2},
    Dongwoo Kim\textsuperscript{\rm 1,2}
}
\affiliations {
    \textsuperscript{\rm 1}CSE, POSTECH\\  
    \textsuperscript{\rm 2}GSAI, POSTECH\\
    \{saemi, shhj1998, dongwookim\}@postech.ac.kr
}

\usepackage{bibentry}

\begin{document}

\maketitle

\begin{abstract}
We tackle the problem of feature unlearning from a pre-trained image generative model: GANs and VAEs. Unlike a common unlearning task where an unlearning target is a subset of the training set, we aim to unlearn a specific feature, such as hairstyle from facial images, from the pre-trained generative models. 
As the target feature is only presented in a local region of an image, unlearning the entire image from the pre-trained model may result in losing other details in the remaining region of the image. 
To specify which features to unlearn, we collect randomly generated images that contain the target features.
We then identify a latent representation corresponding to the target feature and then use the representation to fine-tune the pre-trained model.
Through experiments on MNIST, CelebA, and FFHQ datasets, we show that target features are successfully removed while keeping the fidelity of the original models. Further experiments with an adversarial attack show that the unlearned model is more robust under the presence of malicious parties.
\end{abstract}

\section{Introduction}



Recent advancements in deep generative models have led to the generation of highly realistic images. However, this progress has also raised concerns about the potential misuse of such models. In some instances, generated images may contain violent or explicit content
and inadvertently leak private information used to train the model. To address these issues, a well-prepared dataset with appropriate cleansing procedures can mitigate the potential for abuse of generative models.

In addition to data preparation and cleansing, machine unlearning serves as a complementary tool for preventing the problem in the development and deployment of a generative model. 
Machine unlearning aims to erase the target data from a pre-trained machine-learning model, which can be required to remove private information, harmful content, and biased information~\cite{cao2015towards}. 
However, most of the machine unlearning methods have been focused on supervised models so far~\cite{gupta2021adaptive, tarun2021fast, baumhauer2022machine, ginart2019making, yoon2022few, chundawat2022zero, golatkar2020eternal, nguyen2020variational}.

\begin{figure}[t!]
\centering
    \includegraphics[width=\linewidth]{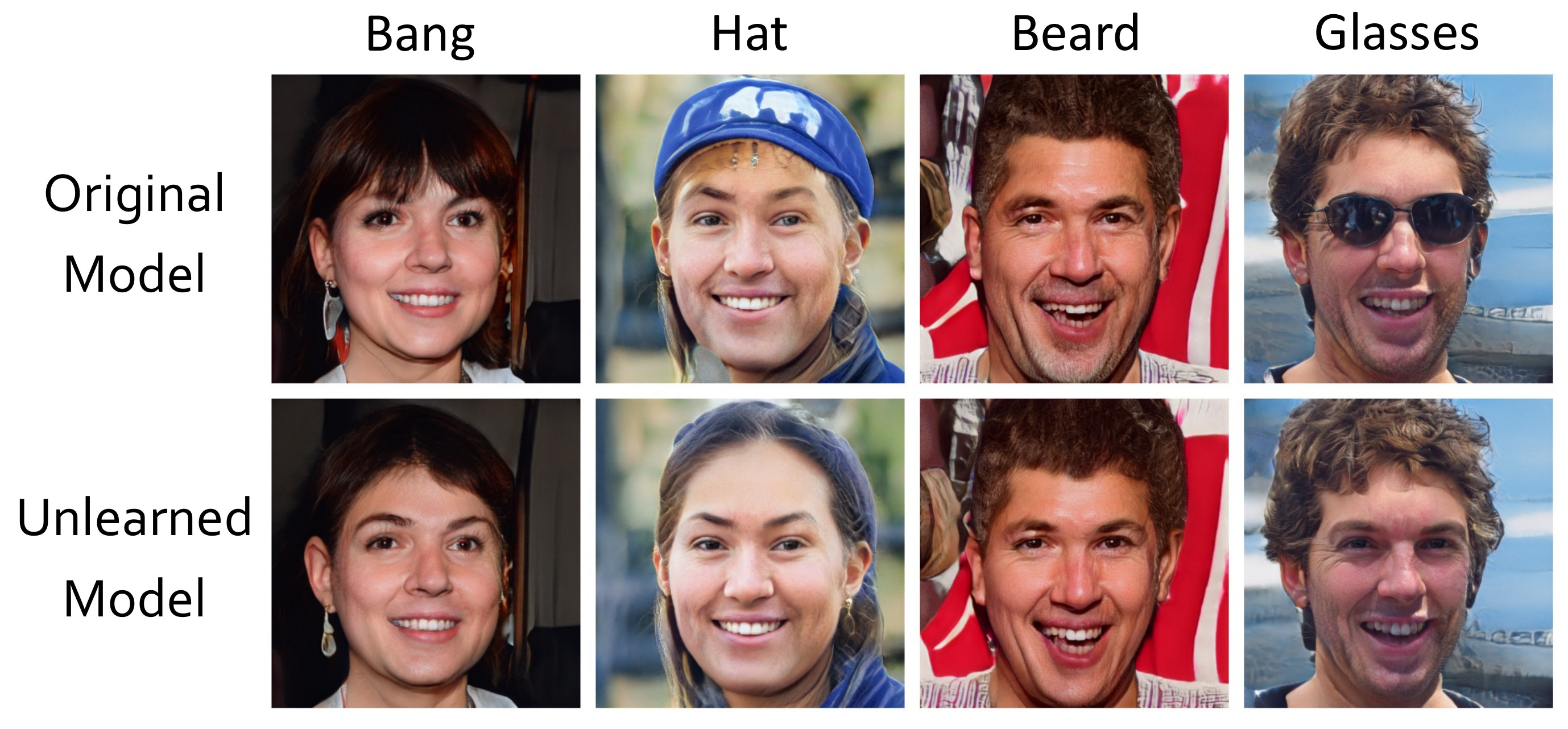}
    \label{fig:celeba_HQ}
    \caption{Result of unlearning various features from pre-trained StyleGAN model. We utilize the same latent vector to generate images from both the original and the unlearned models. Our method effectively unlearns the target feature while maintaining high image quality.}
\label{fig:celebaHQ}
\end{figure}

In this work, we tackle the problem of feature unlearning from pre-trained image generative models where we aim to fine-tune the model to exclude the production of samples that exhibit target features. 
One of the challenges in feature unlearning is that the target feature can be subtle. For instance, a specific hairstyle of a facial image could be the target feature we want to remove from the model, as shown in Figure~\ref{fig:celebaHQ}. 
The subtlety in the target features makes it hard to adopt traditional supervised model unlearning approaches.

In many unlearning scenarios with supervised models, the target of unlearning is a subset of a training dataset, leading to the oracle model that could have been obtained from training without the target subset. Unlike supervised model unlearning, it is non-trivial to define the target data when unlearning feature since the target feature is only presented in a local region of an image.
If we naively remove the entire image that contains the target feature, we could lose the other information in the remaining region of the image. Eventually, subset removal results in a loss of high fidelity and diversity in the generated samples. On the other hand, explicit pixel-level supervision could be given to unlearn the target feature, but it is often very expensive to obtain such supervision.
Furthermore, the problem becomes more challenging if the training dataset is inaccessible during unlearning for several reasons, e.g., storage capacity, private content protection, etc.

To overcome such challenges, we propose a novel generative model unlearning framework that can be applied to generative adversarial networks (GANs) and variational auto-encoders (VAEs). 
To do so, we first collect randomly generated images that contain the target features. We then identify the latent representation of the target features and use the representation to fine-tune the pre-trained model, which prevents the model from generating images with the target feature.
To our knowledge, this is the first framework for unlearning target features in the pre-trained GANs and VAEs. Experimental results show that our unlearning method effectively removes the target feature while maintaining the image quality. 
An additional study based on adversarial attacks also confirms that the proposed method is more robust than a standard model against malicious behavior.
\section{Related Work}

\subsection{Machine Unlearning}

Previous studies have demonstrated that machine learning models may leak sensitive information through attacks or specific inputs~\citep{yuan2019adversarial, cao2015towards}. In addition, regulations have emerged to protect private information, such as `the right to be forgotten', which grants users the request that their personal information must be removed from a system~\citep{rosen2011right}. These highlight the growing significance of machine unlearning.

Unlearning scenarios can vary depending on the requirements~\citep{nguyen2022survey}. Traditional machine unlearning approaches assume that all training data can be accessed~\citep{gupta2021adaptive, tarun2021fast, baumhauer2022machine, ginart2019making}. However, recent studies have presented problem formulations in which access to the data is highly restricted~\citep{yoon2022few, chundawat2022zero, golatkar2020eternal, nguyen2020variational}. In the context of feature unlearning, \citet{guo2022efficient} proposes a representation detachment approach to unlearn the specific attribute for the image classification task. However, the above researches focus on supervised learning tasks, whereas we focus on unsupervised generative models.

Recent research has focused on unlearning methods into generative models. \citet{kong2022data} proposed a data redaction method from pre-trained GAN. They use a data augmentation-based algorithm to prevent making undesirable samples. This method can only be applied when the entire dataset is available. Besides that, we first propose the generative model feature unlearning framework when access to entire data is infeasible. Additionally, \citet{gandikota2023erasing} and \citet{zhang2023forget} propose unlearning methods applicable to text-to-image diffusion models. However, these methods are limited to cross-attention-based models, which may hinder their generalization to diverse generative models. In contrast, our unlearning method can be applied to any generative model that has its own latent space.

\subsection{Latent Space Analysis}
It is known that generative models, such as GANs~\citep{goodfellow2020generative, radford2015unsupervised, karras2017progressive} and VAEs~\citep{kingma2013auto, child2020very} well preserve the information of data within a low-dimensional space, referred to the latent space. In recent years, various techniques for traversing the latent space and extracting a latent vector that represents a visual feature have been proposed.

\citet{radford2015unsupervised} obtain the visual feature vector by subtracting the two latent vectors: the mean latent of the images without the features and the mean latent of the images with the features.
To decide the label of a given latent vector of an image, several approaches attempt to learn a predictor with latent vectors labeled with the corresponding features
~\citep{goetschalckx2019ganalyze, tran2017disentangled,shen2020interpreting}. Unsupervised methods for finding interpretable axes in the generator have also been proposed~\citep{harkonen2020ganspace, voynov2020unsupervised, tzelepis2021warpedganspace, shen2021closed, wang2021geometry}.

In our proposed framework, obtaining the target vector representing the target feature in the latent space is a crucial step. We introduce a straightforward and user-friendly approach that can be applied to both GANs and VAEs, which can be applied to real-world scenarios easily. Additionally, we leverage the target vector to the target identification method within latent space.


\section{Feature Unlearning for Generative Models}
In this section, we propose a framework for unlearning generative models such as GANs and VAEs to make the unlearned model unable to generate the target feature. Throughout this work, we assume that the training dataset is inaccessible once the training is done due to various reasons, such as limited storage or privacy concern.

\subsection{Dataset Preparation}
\label{sec:feature}

\begin{figure}[t!]
    \centering
    \includegraphics[width=\linewidth]{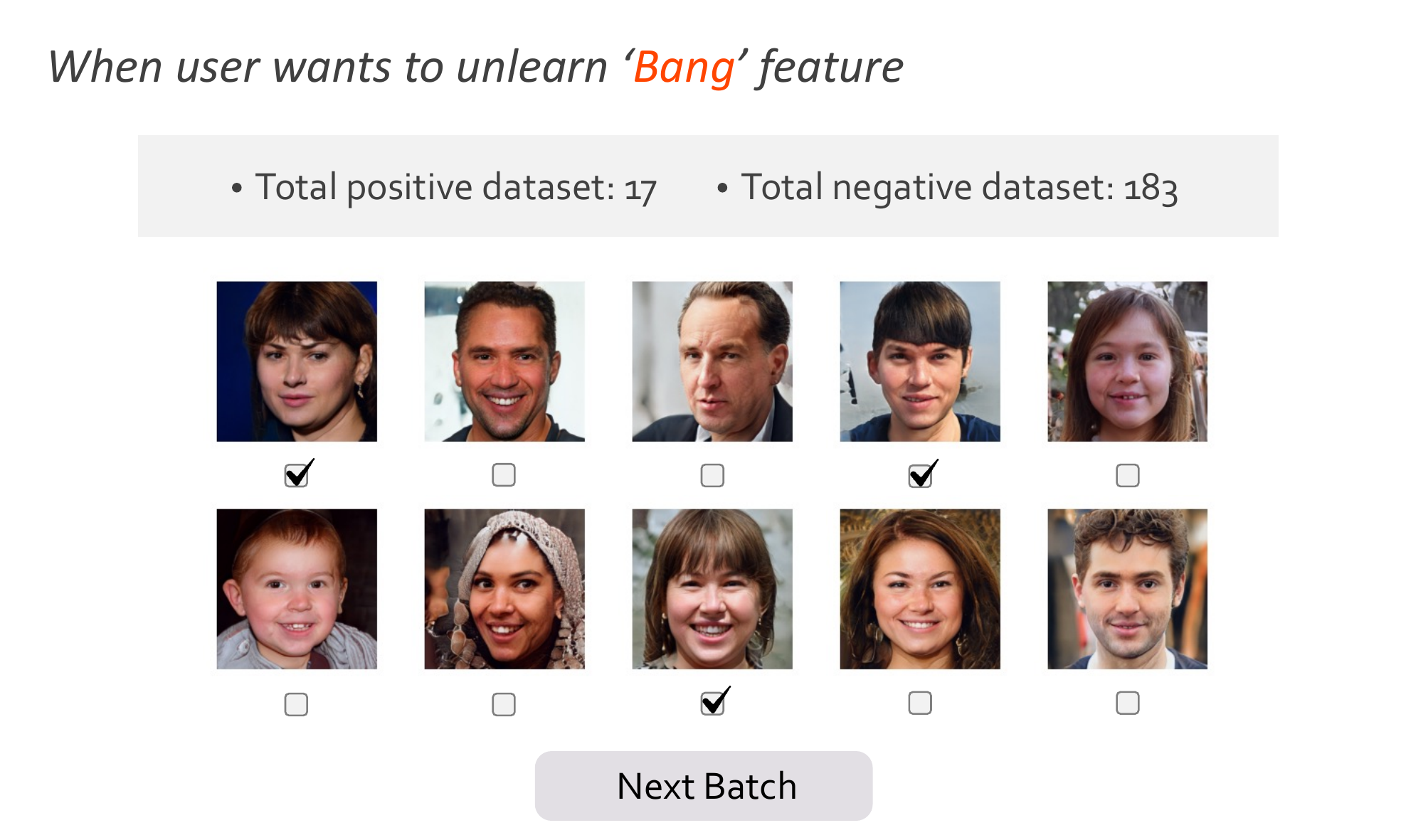}
    \caption{Illustration of interface used to collect images containing the target feature from generated images. A user selects images that contain the target feature to be unlearned. The selected and non-selected images serve as positive and negative datasets for target feature identification.}
\label{fig:select}
\end{figure}

Feature unlearning aims to remove a specific feature from a pre-trained generative model. For example, after unlearning the smile feature from a generative model trained on the CelebA dataset, the model would never generate images of a smiling person.
To do so, in our framework, we first collect datasets using the principle of distance supervision~\cite{mintz2009distant}. 
We curate a `positive' dataset with images that contain the feature to be erased from generated images. The rest of the images without the target feature is categorized into a `negative' dataset. In practice, we can develop an interface where users can select images that contain the target feature. Figure \ref{fig:select} displays the prototype system that we develop to collect the user responses. 

    %


\subsection{Unlearning Framework}

\begin{figure*}[t!]
\centering
    \includegraphics[width=\linewidth]{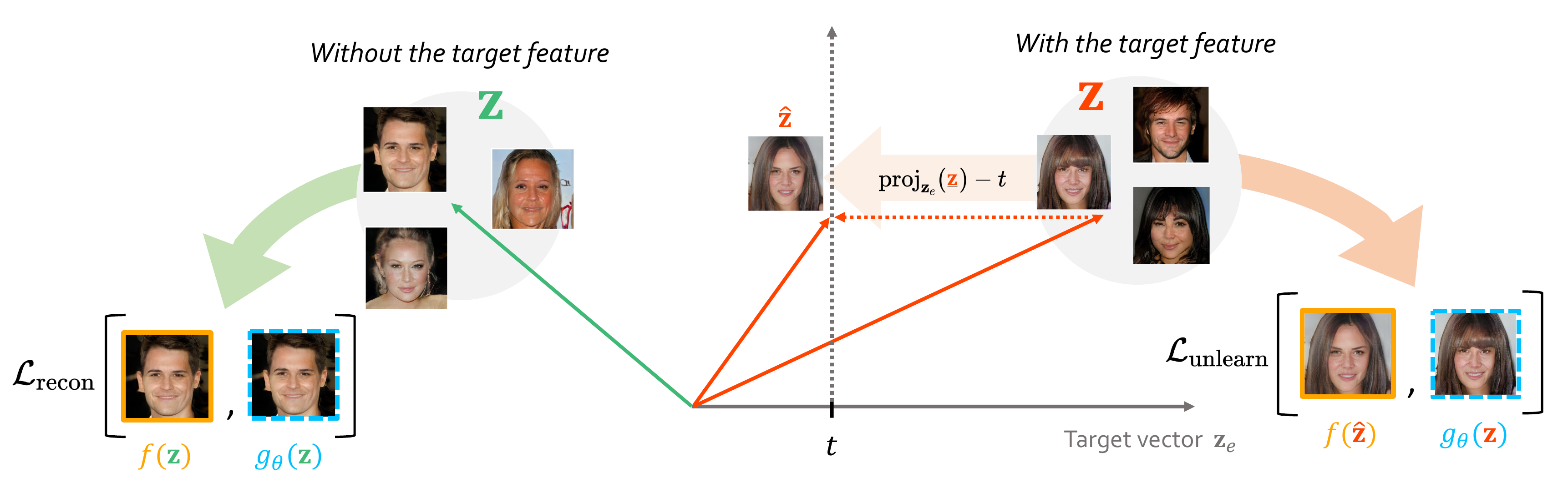}
    \caption{Overall illustration of the generative model unlearning framework. Based on whether the randomly sampled vector $\rvz$ has the target feature, we use different loss functions to unlearn the target feature. $t$ refers to a threshold, and $\hat{\rvz}$ is the translated vector, i.e., $\hat{\rvz} = \rvz - ( \operatorname{proj}_{\rvz_e}(\rvz)  - t)\rvz_e$.}
\label{fig:framework}
\end{figure*}
Unlearning the entire subset with the target feature may lead to losing other image details. Pixel-level supervision can specify areas to unlearn, but it is expensive to scale. Instead, we unlearn the target feature by learning a transformation from the image containing the target feature to the image without the one. To learn such transformation, we need a paired dataset with and without target features. For example, if the target image represents a man smiling and wearing a hat, and the target feature is the smile, we need the non-target image that depicts the same man with the hat but without the smile. However, it is impossible to curate such a dataset in a real-world scenario.

Since our goal is to unlearn the pre-trained generator and not to learn the transformation, we need to apply the transformation principle to the sampling process of the generator. A transformation in image space can be modeled by the corresponding transformation in the latent space.

Based on this intuition, we propose a general unlearning framework from the latent variable perspective as follows:
 \begin{enumerate}
\item Collect positive and negative datasets from generated images.
\item Find a latent representation $\rvz_e$ that represents the target feature in the latent space.
\item \label{step} Sample a latent vector $\rvz$ from a simple distribution.
 \begin{enumerate}
\item \label{step_recon} If the latent vector does not contain the target feature, let the generator produce the same output without modification.
\item \label{step_scrub} If the latent vector contains the target feature, fine-tune the generator to produce a transformed output without the target feature.
\end{enumerate}
\item Repeat step \ref{step} until the generator does not produce the target feature.
\end{enumerate}

\noindent\textbf{Target identification in latent space.} 
\label{par:latent_representation}
We assume that a vector in the latent space can represent the target feature. As the first step of unlearning, we obtain the latent vector representation of each image from the collected dataset.
Once we obtain the latent vectors, we use a vector arithmetic method proposed by \citet{radford2015unsupervised} to find the latent vector representing the target feature. Specifically, we compute the mean vectors from a collection of a positive dataset and a negative dataset and subtract the mean vectors of the negative dataset from that of the positive dataset. The resulting \emph{target vector} $\rvz_e$ is then used to represent the target feature in the latent space.

To determine whether a randomly generated image contains a target feature, we project its latent vector onto the target vector. \citet{white2016sampling} shows that the projection can represent the similarity between the latent vector and the target feature. We then compare this value to a threshold to determine whether the image contains the target feature. For the experiments, we set the threshold value $t$ as the average projection values of the positive and negative samples in the latent space. Let $\operatorname{sim}(\rvz, \rvz_e) \in \{0,1\}$ indicate the binary classification results, i.e.,
\begin{align}
\label{eqn:feature_cls}
    \operatorname{sim}(\rvz, \rvz_e) = \begin{dcases*}
        0, & if $  \operatorname{proj}_{\rvz_e}(\rvz)  < t$,\\
        1, & otherwise,\\
        \end{dcases*}
\end{align}
where $\operatorname{proj}_{\rvz_e}(\rvz)$ is the projection of $\rvz$  onto $\rvz_e$, i.e., $\frac{\rvz_e^\top \rvz}{\lVert\rvz_e\rVert}$.

\smallskip\noindent\textbf{Unlearning process.} 
To formalize the unlearning process, let ${g}_\theta$ be the model to be unlearned, and $f$ be the pre-trained generator. We initialize ${g}_\theta$ from the pre-trained $f$. 
When the randomly sampled latent vector $\rvz$ does not contain the target feature, i.e., $\operatorname{sim}(\rvz, \rvz_e)=0$, the generator $g_\theta$ needs to produce the same output as $f$, c.f, step 3 (a). To enforce the minimal changes in the produced output, we formulate the following \emph{reconstruction} objective to minimize
\begin{align}
\label{eq:loss_reconstruction}
\mathcal{L}_{\text{recon}}(\theta) = (1 - \operatorname{sim}(\rvz, \rvz_e)) \lVert {g}_{\theta}(\rvz) - {f}(\rvz) \rVert_1,
\end{align}
where $\rvz$ is the random vector. Hence, the unlearned model $g_\theta$ tries to mimic the original generator when the latent vector does not contain the target feature. 

When the randomly sampled vector contains the target feature, the generation process needs to be changed such that the sampled output no longer contains the target feature.
To do so, we first create the target-erased output by generating an output with a translated random vector using $f$. Given random vector $\rvz$, we first project the vector onto the target vector, and then the original random vector is shifted by the projected vector, i.e., $\rvz - (\lVert \operatorname{proj}_{\rvz_e}(\rvz) \rVert - t)\rvz_e$, where $t$ is the predefined threshold. The translated vector is used as an input of the original generator $f$ producing the target-erased output. The modified output is then used to train $g$ with the following \emph{unlearning} objective 
  \begin{align}
    \label{eq:loss_unlearn}
    \begin{split}
    \mathcal{L}_{\text{unlearn}}
    &(\theta) = \operatorname{sim}(\rvz, \rvz_e)  \\
    &\lVert {g}_{\theta}(\rvz)-{f}\left(\rvz- \left( \operatorname{proj}_{\rvz_e}(\rvz) -{t}\right)\rvz_e\right) \rVert_1\;.
    \end{split}
  \end{align}
The objective enforces the unlearned generator producing outputs similar to those from the original generator without target features.
If the projection can correctly measure the presence of the target feature in the latent space while disentangling the other features, $g_\theta$ can successfully forget the target feature in the latent space.

It is widely known that L2 and L1 loss occurs in blurry effects in image generation and restoration tasks~\citep{pathak2016context, zhang2016colorful, isola2017image, zhao2016loss}. Prior research has addressed the blurry effects by introducing diverse techniques, such as adding perceptual or adversarial loss to the training process~\cite{johnson2016perceptual, zhao2016loss}. 
To overcome the blurry effects, we add \emph{perceptual} loss into the objective function. The objective is formalized as
\begin{align}
    \begin{split}
    \label{eq:loss_perception}
        &\mathcal{L}_{\text{percep}}(\theta) = \operatorname{sim}(\rvz, \rvz_e)\\
        & \left(1-\operatorname{MS-SSIM}\left({g}_{\theta}(\rvz),{f}\left(\rvz- \left( \operatorname{proj}_{\rvz_e}(\rvz) -{t}\right)\rvz_e\right)\right)\right),
    \end{split}
\end{align}
where MS-SSIM function refers to the Multi-Scale Structural Similarity~\cite{zhao2016loss}, which measures perceptual similarity between two images by comparing luminance, contrast, and structural information.

Finally, we combine three objective functions to define the unlearning objective as
\begin{align}
    \label{eq:loss_total}
    \mathcal{L}(\theta) = \alpha \left( \mathcal{L}_{\text{unlearn}}(\theta) + \mathcal{L}_{\text{percep}}(\theta)\right) + \mathcal{L}_{\text{recon}}(\theta)\;,
\end{align}
where $\alpha$ is the hyper-parameter that regulates the unlearning and reconstruction error balance. We visualize the overall framework in Figure \ref{fig:framework}. 





\section{Experiments}

\begin{table*}[t!]
\centering
\resizebox{\textwidth}{!}{%
\begin{small}
\begin{tabular}{cccrrrrrrrrr}
\toprule
\multirow{2}{*}{Dataset} & \multirow{2}{*}{Model} & \multirow{2}{*}{Feature} & \multicolumn{3}{c}{Target feature ratio (\%)}                                                          & \multicolumn{3}{c}{Inception Score}                                                                    & \multicolumn{3}{c}{Fréchet Inception Distance}                                                                   \\ \cmidrule{4-12} 
                         &                        &                          & \multicolumn{1}{c}{Original} & \multicolumn{1}{c}{Unlearn} & \multicolumn{1}{c}{Oracle} & \multicolumn{1}{c}{Original} & \multicolumn{1}{c}{Unlearn} & \multicolumn{1}{c}{Oracle} & \multicolumn{1}{c}{Original} & \multicolumn{1}{c}{Unlearn} & \multicolumn{1}{c}{Oracle} \\ \midrule
\multirow{4}{*}{MNIST}   & \multirow{2}{*}{VAE}   & Thickness                & 9.47                       & 0.97                      & 1.01                       & 2.21                        & 2.16                       & 2.14                        & 23.36                       & 23.74                      & 24.03                       \\
                         &                        & Slant                    & 9.22                       & 0.91                      & 1.08                       & 2.19                        & 2.22                       & 2.22                        & 22.84                       & 23.27                      & 23.52                       \\
                         & \multirow{2}{*}{DCGAN} & Thickness                & 9.10                       & 1.35                      & 1.28                       & 2.14                        & 2.13                       & 2.11                        & 2.32                        & 3.35                       & 3.40                        \\
                         &                        & Slant                    & 10.92                       & 1.25                      & 1.31                       & 2.15                        & 2.14                       & 2.10                        & 2.24                        & 2.79                       & 2.85                        \\ \midrule
\multirow{4}{*}{CelebA}  & \multirow{2}{*}{VDVAE} & Bang                     &  3.36                            &         0.28                    &                0.21             &     2.57          &      2.58             &          2.54          &        82.92           &   85.21                  &        84.09             \\
                         &                        & Beard                    &      7.22                        &           2.41                  &   1.09                           &     2.47        &    2.38                  &         2.39             &                82.92     &           86.15        &            84.49        \\
                         & \multirow{2}{*}{ProgGAN} & Bang                     & 6.74                       & 0.42                      & 0.49                       & 2.92                        & 2.91                       & 2.91                        & 48.05                       & 49.82                      & 51.37                       \\
                         &                        & Beard                    & 3.02                       & 1.01                      & 0.98                       & 2.93                        & 2.88                       & 2.87                        & 48.05                       & 49.80                      & 49.78                       \\ \midrule
\multirow{4}{*}{FFHQ}& \multirow{4}{*}{StyleGAN} & Bang                     &  4.48                            &         0.26                    &               \ding{55}            &     3.61          &      3.33             &          \ding{55}          &        20.97           &   25.88                  &       \ding{55}             \\
                         &                        & Beard                    &      21.96                        &           1.37                  &  \ding{55}                           &     3.60        &    3.34                  &         \ding{55}            &                20.93     &           25.12        &           \ding{55}        \\  &&Hat                     &  2.10                            &         0.12                    &               \ding{55}            &     3.62          &      3.37             &          \ding{55}          &        21.00           &   24.75                  &       \ding{55}             \\
                         &                        & Glasses                    &      6.16                        &           0.19                  &  \ding{55}                           &     3.61        &    2.37                  &         \ding{55}            &                20.86     &           23.86        &           \ding{55}\\                    
                         \bottomrule

\end{tabular}
\end{small}
}
\caption{Target feature ratio ($\downarrow$), inception score ($\uparrow$), and Fréchet inception distance ($\downarrow$) of original, unlearn, and oracle models.}
\label{table/result}
\end{table*}

In this section, we show the performance of the proposed framework for unlearning GANs and VAEs trained on three datasets. We conduct both synthetic experiments and user studies with human participants.


\subsection{Experiment Setup}
\label{subsec:experimental_setting}
\noindent\textbf{Datasets and models.}
To show the performance of the proposed framework, we conduct experiments on three datasets with a different set of generative models for each. We list the three datasets used in the experiments and the generative models trained on each dataset as follows:
\begin{itemize}
    \item \textbf{MNIST}~\cite{lecun1998gradient}: a Deep Convolutional GAN (DCGAN)\cite{radford2015unsupervised} and a vanilla VAE~\cite{kingma2013auto}
    \item \textbf{CelebA}~\cite{liu2018large}: Progressive Growing of GANs (ProgGAN)~\cite{karras2017progressive} and a Very Deep VAE (VDVAE)\footnote{We concatenate multiple layers of latent variables to apply the unlearning algorithm.}~\cite{child2020very} 
    \item \textbf{FFHQ}~\cite{karras2019style}: a StyleGAN\footnote{We use the output of the mapping network to unlearn.}~\cite{karras2019style}
\end{itemize}
Further details of the models can be found in Section \ref{subsec:model_detail}.

\smallskip\noindent\textbf{Unlearning dataset preparation.}
To understand how the proposed algorithm works, we simulate the cases by using the known features of each dataset. Specifically, we select two features from each dataset, with each feature representing approximately 10\% of the dataset.

For the MNIST dataset, we choose the thickness and left slant as target features to unlearn. We use the Morpho-MNIST~\cite{castro2019morphomnist}, which provides a comprehensive tool for measuring various features of MNIST digits. Since the thickness and left slant are not binary, we use images whose feature values range within the top 10\% of the entire dataset as a positive dataset and the remaining as a negative dataset. The CelebA provides 40 features for each image. Among the available features, we choose `Bang' and `Beard' as the target features to be unlearned. Although FFHQ does not provide annotations for each image, we decided to unlearn the same features as CelebA since both are facial datasets. We experiment with two additional features, `Hat' and `Glasses' for FFHQ.

For this experiment, we use a classifier to categorize each sampled image into a positive or negative dataset. To get a target feature, we utilize Morpho-MNIST as the classifier and categorize 500 generated images for MNIST. For both CelebA and FFHQ, we use a pre-trained MobileNet~\citep{howard2017mobilenets} to classify 5,000 generated images. We use these classifiers as a proxy of users, but we also conduct a user study involving human participants reported in Section \ref{sec:user_study}.

\smallskip\noindent\textbf{Oracle model.}
There is no unlearning method targeted for generative models, according to our information. To evaluate effectiveness, we compared our framework with the oracle model trained from scratch without the images containing the target feature. The oracle model is commonly used as a standard oracle model in supervised unlearning cases, whereas in our case, this is not ideal since the positive dataset can contain useful features other than the target feature. 
Note that, for FFHQ dataset, we cannot train the oracle model since the annotated features are not available.

\smallskip\noindent\textbf{Training details.}
For all experiments, we use Adam optimizer and a learning rate of 0.001, 0.002, and 0.005 for MNIST, CelebA, and FFHQ respectively. The MNIST dataset is trained for 200 epochs, and CelebA and FFHQ are trained for 500 epochs in unlearning process. We use NVIDIA GeForce RTX 3090 and A6000 for experiments.

\subsection{Evaluation Metric}

The performance of unlearning can be measured from two different perspectives: 1) how well the unlearning is done and 2) how good the sample qualities are. We explain two different metrics used to evaluate the models.

\smallskip\noindent\textbf{Target feature ratio.}
\label{par:target_feature_ratio}
The target ratio measures the percentage of generated samples with the target feature. Low values of the target ratio indicate that the generative model has successfully unlearned the target feature. We use the same pre-trained classifiers used to target feature identification. For all experiments, we randomly sample 10,000 images from a generative model to compute the target ratio.

\smallskip\noindent\textbf{Image quality.}
We report two commonly used metrics to evaluate the quality of the generated image: Inception Score (IS)~\cite{salimans2016improved} and Fréchet Inception Distance (FID)~\cite{heusel2017gans}. We compute FID between the generated samples and the original training dataset. Higher IS scores and lower FID scores indicate higher image quality. We use an implementation of StudioGAN~\cite{kang2022studiogan} to calculate IS and FID. 50,000 samples are used to measure the scores. 

\begin{figure*}[t!]
\centering
    \includegraphics[width=.95\linewidth]{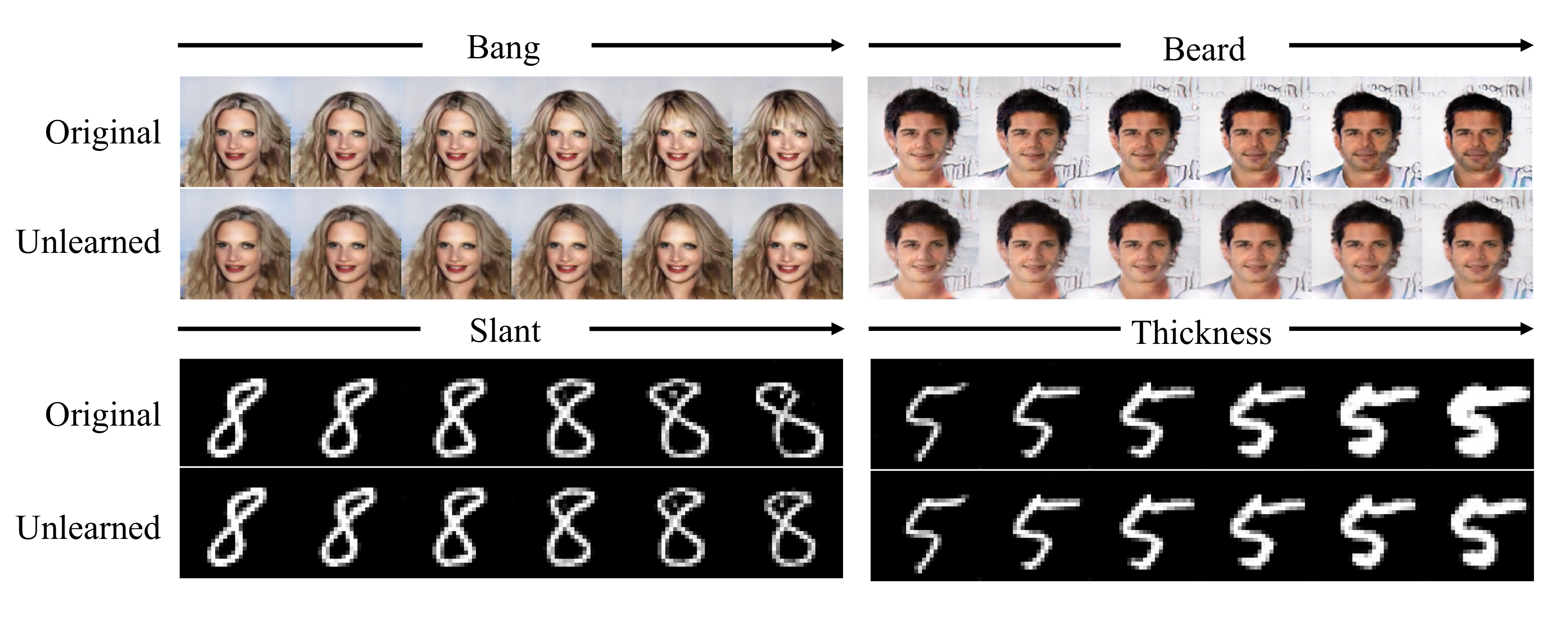}
    \caption{Visualization of four different features before and after unlearning from pre-trained GAN models. All paired images in each column are generated from the same latent vector.}
\label{fig:result}
\end{figure*}

\subsection{Results}

\noindent\textbf{Quantitative results.}
We evaluate the effectiveness of our unlearning framework by comparing the target feature ratio between the original model, the unlearned model, and the oracle model in Table \ref{table/result}. The results show that the unlearned model produces similar target feature ratios to the oracle for all features, indicating our framework successfully unlearns the target feature. 

Ensuring high image quality is also important in unlearning the target feature. Table \ref{table/result} presents the results of the IS and FID scores for evaluating the quality of generated images, respectively. The results demonstrate that all three models produce similar IS and FID scores, indicating our framework can successfully unlearn the target feature while maintaining high-quality image generation. Note that FID is calculated using the entire dataset, which yields a slightly higher FID value for the unlearned and oracle models, but there is no significant difference between the two models.

\smallskip\noindent\textbf{Qualitative results.}
Figure \ref{fig:result} presents the qualitative visualization result of our unlearning framework. The top row shows the images generated from the original generator, and the bottom row shows those generated from the unlearned generator. By visualizing generated images using the same latent vector, we observe that the target feature has been effectively erased in each case. 
In addition, we unlearn various features from StyleGAN trained with FFFQ dataset~\cite{karras2017progressive}, whose resolution is higher than the other two datasets. The qualitative results in Figure \ref{fig:celebaHQ} show the approach also works well with high-resolution images. Additional qualitative results are provided in Section \ref{subsec:add_result}.

\smallskip\noindent\textbf{Computational efficiency.}
Our unlearning framework takes approximately one minute to unlearn MNIST on a single GPU for VAE and DCGAN, and approximately 10 minutes to unlearn CelebA and FFHQ with four GPUs for ProgGAN, VDVAE, and StyleGAN. 
In contrast, training the oracle for MNIST requires approximately 30 minutes on a single GPU. Training the oracle for ProgGAN on CelebA takes around three days using eight GPUs, and one for VDVAE takes about two days using four GPUs. The authors of StyleGAN report that training StyleGAN on FFHQ takes approximately 6 days and 14 hours with 8 Tesla V100 GPUs\footnote{https://github.com/NVlabs/stylegan}. Although we assume that relearning is impossible, nevertheless, even if relearning were possible, our method is significantly more time-efficient with comparable results.



\subsection{User Study}
\label{sec:user_study}

A user study was conducted to assess the effectiveness of our unlearning framework in a more realistic scenario. This study was designed to scrutinize the performance of the unlearned model in comparison to the original model across various dimensions. We recruited 13 participants and asked them to select images that contain `Glasses' since the feature is distinctly discernible by users.
500 samples were annotated by each participant using the interface shown in Figure \ref{fig:select}. The annotation process took roughly 5 minutes on average. Then, we unlearned the pre-trained StyleGAN trained on FFHQ for each participant. User study details and screenshots are provided in Section \ref{sec:user_detail}.

We use the following three criteria to measure the performance of unlearning against the original model:
\begin{enumerate}
    \item \textbf{Target feature ratio.}
This task involved counting the number of target images (images with glasses) among randomly generated images. A participant examined 500 images, each of which are randomly chosen between the original and unlearned models.
\item \textbf{Image quality.}
Participants were asked to choose an image with better quality. We provided two randomly generated samples from the original and unlearned models, one for each. The samples were shuffled before being presented, and each participant evaluated a total of 50 cases. Users can respond that one of the two images is of better quality or that both images are of similar quality.
\item \textbf{Pinpoint unlearning.}
We provided two images generated with the same latent vector from the original and unlearned models and asked participants how many features were changed after unlearning. CelebA's attributes were shown in advance to familiarize participants with the existing features. The participants had the following options:
\begin{itemize}
    \item All features except the target were unchanged.
    \item One or two features appeared to have changed.
    \item More than two features appeared to have changed.
\end{itemize}
Each participant compared ten randomly generated pairs.
\end{enumerate}

\noindent\textbf{Result.}
\begin{figure}[t!]
    \centering
    \includegraphics[width=\linewidth]{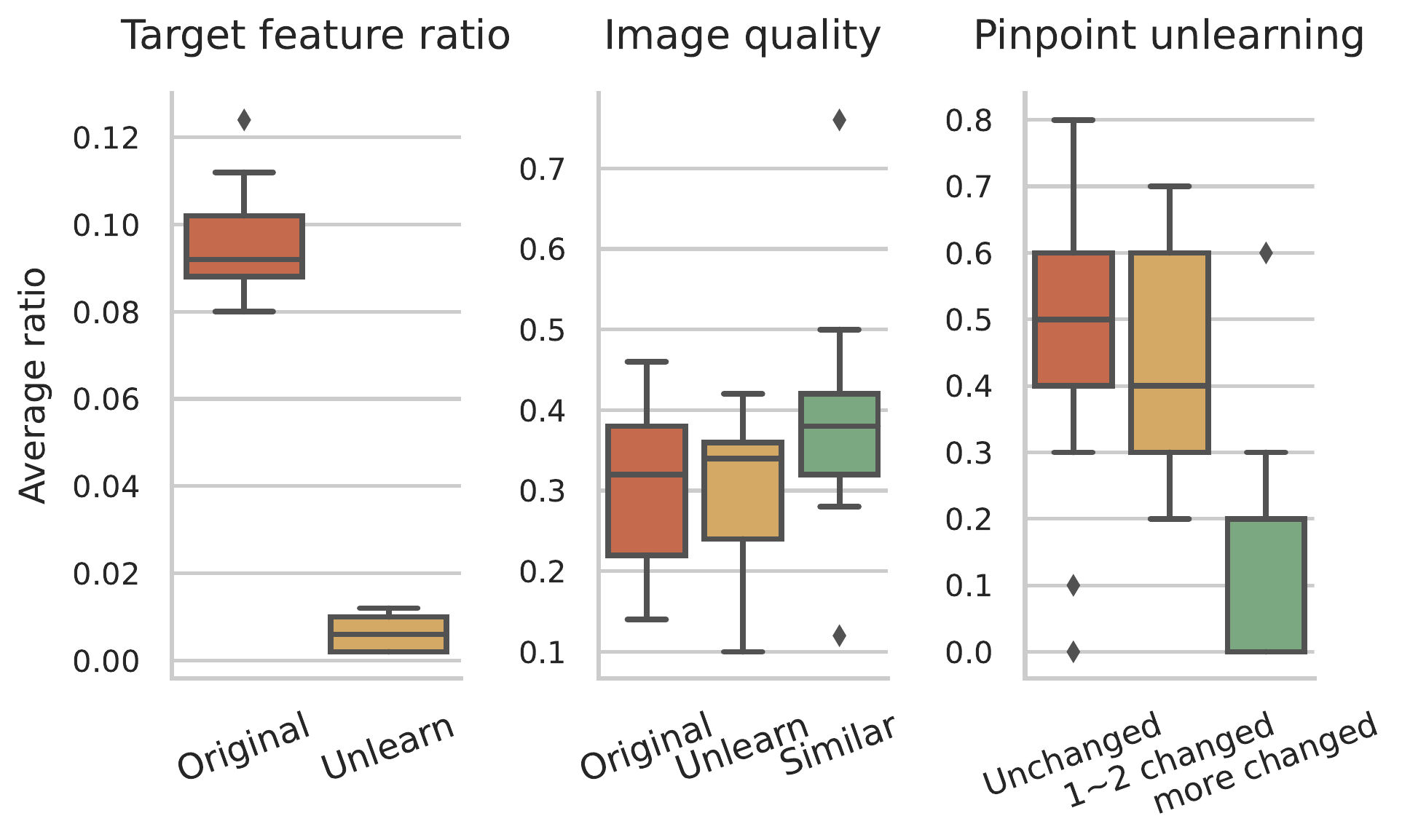}
    \caption{User study result of unlearning `Glasses' feature from the StyleGAN.}
\label{fig:userstudy}
\end{figure}
Figure \ref{fig:userstudy} presents the result of the user study. We normalize the answers of each user and plot their distribution. The result shows a significant drop in the target feature ratio representing the effectiveness of our framework on the target feature. There is no significant difference in image quality between the original and unlearned models, and the majority of users answer the quality between the two models is similar to each other.

Participants evaluated that only the target feature was changed through unlearning on average 45\% of cases. However, 43\% of cases were reported to have one or two features changed, and 12\% to have more than two features changed.
We found that the entanglement between the `Glasses' and `Young' features leads to the result. In FFHQ, a person with `Glasses' are more likely to be elderly. As a consequence, the identified target feature vector is likely to be entangled with the `Young' feature. More sophisticated feature disentanglement algorithms~\cite{tran2017disentangled,locatello2019challenging} could help to mitigate such effect, but we leave this for future work.
\section{Adversarial Attack}
\label{sec:adversarial_attack}

To check the robustness of our unlearning method under the presence of adversaries, we further conduct an experiment with an adversarial attack method on an unlearned model.

\subsection{Experimental Setting}
An adversarial attack on the unlearned model is conducted to assess its vulnerability to malicious users who may attempt to exploit the model to generate harmful or explicit content. By subjecting the unlearned model to an adversarial attack, we can evaluate its robustness and ensure that the unlearned model does not generate content that goes against ethical or safety guidelines.

For this experiment, we employ a Projected Gradient Descent (PGD) attack~\cite{madry2017towards} on the latent variable of the unlearned model. The attack is guided by a pre-trained feature classifier capable of classifying the target feature. The purpose of this attack is to determine whether the unlearned model can be manipulated to produce images that contain the target feature, even after it has been supposedly removed. 
The overall method used in this experiment is provided in Algorithm \ref{alg:attack}.

We attack the ProgGAN~\cite{karras2017progressive} trained with CelebA-HQ~\cite{karras2017progressive} with the target feature of `Bang'. We conduct an adversarial attack on 10,000 distinct latent vectors to both the original and unlearned models.
The classifier used for this experiment is a MobileNet~\cite{howard2017mobilenets} specified in Section \ref{subsec:experimental_setting}. For the hyper-parameters of the PGD method, a learning rate of 0.02 is chosen, and the attack step is set at 50. The maximum magnitude of the permissible perturbation is set to 0.1, i.e., $||\tilde \rvz - \rvz||_\infty \leq 1$. Note that a larger perturbation may cause the perturbed latent vector to deviate significantly from the original distribution.


\subsection{Results}
\begin{table}[t!]
\centering
\begin{small}

\begin{tabular}{ccrr}
\toprule
\multicolumn{2}{c}{}                                                                             & Original & Unlearn \\ \midrule
\multirow{3}{*}{\begin{tabular}[c]{@{}l@{}}Before\\ attack\end{tabular}} & Target feature ratio ($\downarrow$)     &  3.54  & 0.96  \\
                                                                         & IS ($\uparrow$)                   & 3.16     & 3.16  \\
                                                                         & FID ($\downarrow$)                  &  23.54  &  24.00 \\\midrule
\multirow{3}{*}{\begin{tabular}[c]{@{}l@{}}After\\ attack\end{tabular}}  
                                                                         & Target feature ratio ($\downarrow$)&  14.28  &  5.54  \\
                                                                         & IS ($\uparrow$)                   &  2.83   &  3.03  \\ 
                                                                         & FID ($\downarrow$)                  &  44.25  & 38.39 \\\bottomrule
\end{tabular}%
\end{small}
\caption{The target feature ratio (\%)/ IS / FID before and after an adversarial attack.}
\label{table/attack}
\end{table}
As a preliminary step, we evaluate the feature target ratio and image quality for each original and unlearned model to compare the before and after attack results. Table \ref{table/attack} presents the result before the attack. The result shows that the unlearned model generates less number of target images than the original model while maintaining high image quality in terms of IS and FID.

To measure the target feature ratio after the adversarial attack, we train the second target feature classifier with a different random seed. Since the adversarial attack perturbs the latent vector to fool the target classifier, the perturbed sample may not contain the target feature but overfit the target classifier. Hence, we use the second target classifier to measure the target feature ratio.
Although the target feature ratio of the unlearned model has increased after the attack, the target feature ratio of the unlearned model remains lower than that of the original model.

\begin{figure}[t!]
\centering
    \includegraphics[width=\linewidth]{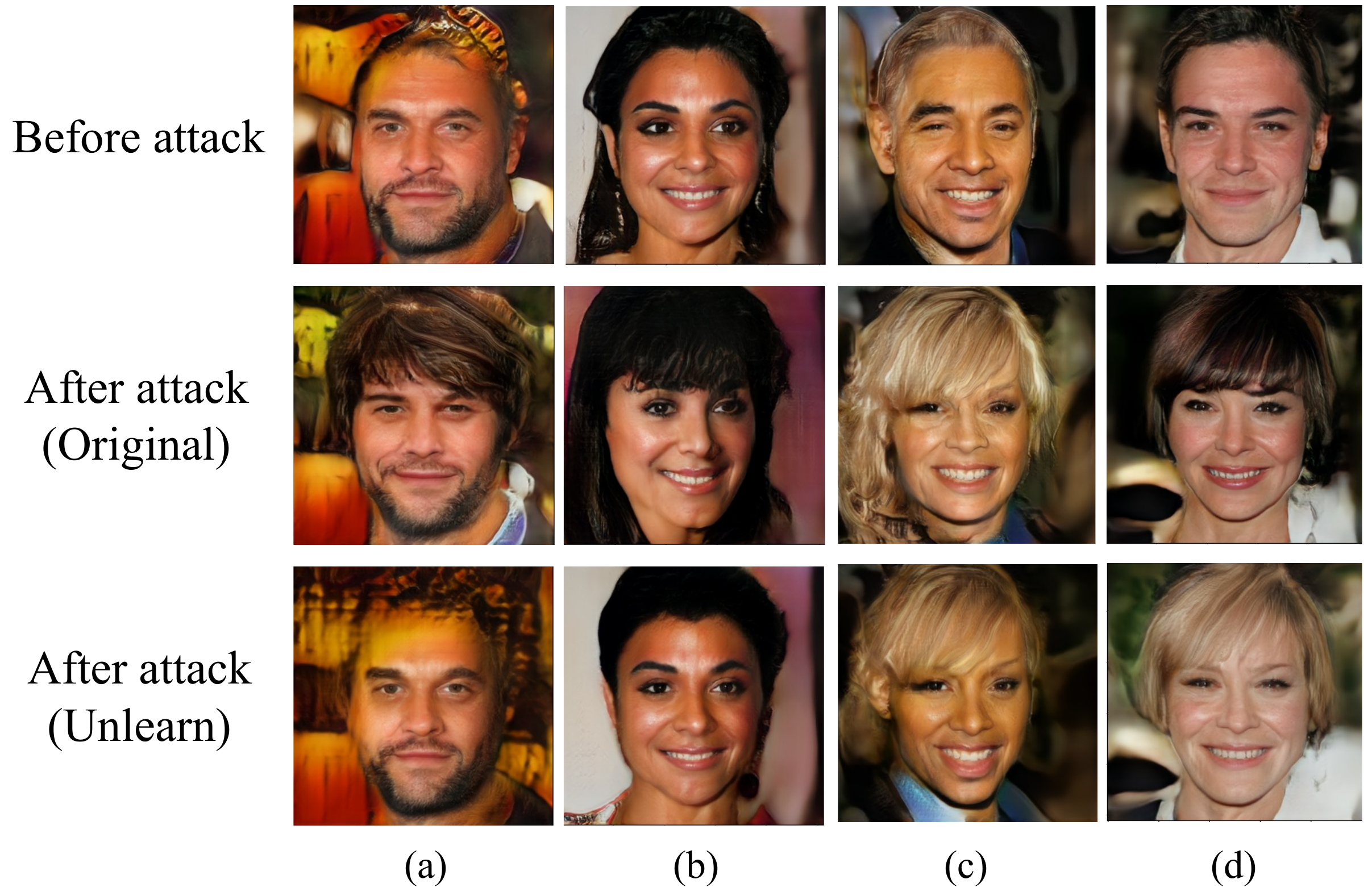}
    \caption{Examples of generated images before and after an adversarial attack on its latent variable. The top row shows the original image. 2nd and 3rd rows are attacked images from the original and unlearned models. The first two columns are successful defenses by the unlearned model and the last two show failures.}
\label{fig:attack}
\end{figure}

We visualize the randomly generated images before and after an adversarial attack in Figure \ref{fig:attack}. The first two columns (a, b) show that the original model generates the target feature through the adversarial attack, while the unlearned model effectively defends against the attack. However, the last two columns (c, d) demonstrate that both models fail to defend against the attack. 

\section{Conclusion}
The recent success of generative models has brought exciting developments in various fields, such as computer vision, natural language processing, and art generation. However, the potential risks associated with the generation of harmful or private content through these models highlight the importance of developing effective unlearning algorithms. Our proposed unlearning algorithm for generative models shows promising results in preventing the generation of unwanted features, which can serve as a crucial tool in addressing sensitive or private content concerns. Future research can build upon this work to improve the efficiency and effectiveness of unlearning algorithms in other contexts, such as data privacy and fairness. Ultimately, the development of robust and reliable unlearning algorithms can maximize the benefits of generative models while minimizing the associated risks.

\section*{Acknowledgements}
This work was partly supported by Institute of Information \& communications Technology Planning \& Evaluation (IITP) grant funded by the Korea government (MSIT) (No.2019-0-01906, Artificial Intelligence Graduate School Program (POSTECH)) and National Research Foundation of Korea (NRF) 
grant funded by the Korea government (MSIT) (No. RS-2023-00217286) and National Research Foundation of Korea (NRF) grant funded by the Korea government (MSIT) (NRF-2021R1C1C1011375). Dongwoo Kim is the corresponding author.

\bibliography{aaai24}

\newpage

\appendix
\onecolumn
\newpage

\section*{Appendix}
We discuss the model details, training details, and hyper-parameter setting in Section~\ref{sec:experiment_detail}. we present additional qualitative results and further analysis of experiments in Section~\ref{sec:add_result}. We then describe the details of the user study in Section~\ref{sec:user_detail}. Finally, Section~\ref{sec:PGD_attack} offers an expanded explanation of the PGD attack used in Section~\ref{sec:adversarial_attack}.

\section{Experimental Details}
\label{sec:experiment_detail}
\subsection{Model Details}
\label{subsec:model_detail}
\paragraph{MNIST.}
We train DCGAN\footnote{\url{https://github.com/pytorch/examples/tree/main/dcgan}} on the MNIST dataset for 100 epochs with a learning rate of 0.0002. For the VAE, we employ three fully connected layers in both the encoder and decoder, training them for 100 epochs at a learning rate of 0.001.

\paragraph{CelebA.}
We utilize the ProgGAN\footnote{\url{https://github.com/facebookresearch/pytorch_GAN_zoo/}} and train it on the CelebA dataset for 96,000 iterations, employing a learning rate of 0.001. We train the VDVAE\footnote{\url{https://github.com/openai/vdvae}} with the imagenet32 configuration from the code.

\paragraph{FFHQ.}
We use the pre-trained StyleGAN\footnote{\url{https://github.com/rosinality/style-based-gan-pytorch}} model at a resolution of 1024 pixels. The target feature vector is identified within the $W$ space, i.e., the output space of the mapping network. Also, we do not update the mapping function, which maps $z$ space to $W$ space.

\subsection{Training Details}
\paragraph{Target feature of the dataset.}
We specify the number of images that have a target feature within each dataset in Table~\ref{table/celeba}. For MNIST, we set it as having the target feature if it has a measured value using Morpho-MNIST in the top 10\%. Also, we select `Bang' and `Beard' as target features. These features appear around 10\% of the CelebA dataset.
\begin{table}[h!]
\centering
\begin{tabular}{ccrr}
\toprule
                        &           & \multicolumn{1}{c}{Target image} & \multicolumn{1}{c}{Non-target image} \\ \midrule
\multirow{2}{*}{MNIST}  & Thickness & 6,000                            & 54,000                               \\
                        & Slant     & 6,000                            & 54,000                               \\ \midrule
\multirow{2}{*}{CelebA} & Bang     & 30,709                           & 171,890                              \\
                        & Beard     & 33,441                           & 169,158                              \\ \bottomrule
\end{tabular}
\caption{Number of the target image and non-target image for each feature used in the experiments.}
\label{table/celeba}
\end{table}

\paragraph{Oracle model.}
The oracle model is initialized with the trained model. We further train the oracle model without the target features over 50 epochs and 96,000 iterations for MNIST and CelebA, respectively, to remove the target features as done in \citep{nguyen2020variational}.

\subsection{Hyper-parameter Setting}
We detail the hyper-parameter used for each dataset in Table~\ref{tab:hyperparameter}. The same hyperparameters were applied when unlearning the GAN and VAE trained on the same dataset.
\begin{table}[h!]
\centering
\begin{tabular}{crrrr}
\toprule
       & \multicolumn{1}{c}{Learning-rate} & \multicolumn{1}{c}{$\alpha$} & \multicolumn{1}{c}{Epoch} & \multicolumn{1}{c}{\# of sample} \\ \midrule
MNIST  & 0.0001                            & 3                         & 200                       & 500                              \\
CelebA & 0.0002                            & 300                       & 500                       & 500                              \\
FFHQ   & 0.0005                            & 300                       & 500                       & 20                               \\ \bottomrule
\end{tabular}
\caption{Hyper-parameter setting used for each dataset in experiments.}
\label{tab:hyperparameter}
\end{table}

\section{Additional Results}
\label{sec:add_result}
\subsection{Additional Qualitative Results}
\label{subsec:add_result}
We further present qualitative results from StyleGAN, visualizing the interpolation of four features: Bang, Beard, Hat, and Glasses. Our results indicate that the target features are removed while preserving high image quality.
\begin{figure}[h!]
    \centering
    \includegraphics[width=0.76\linewidth]{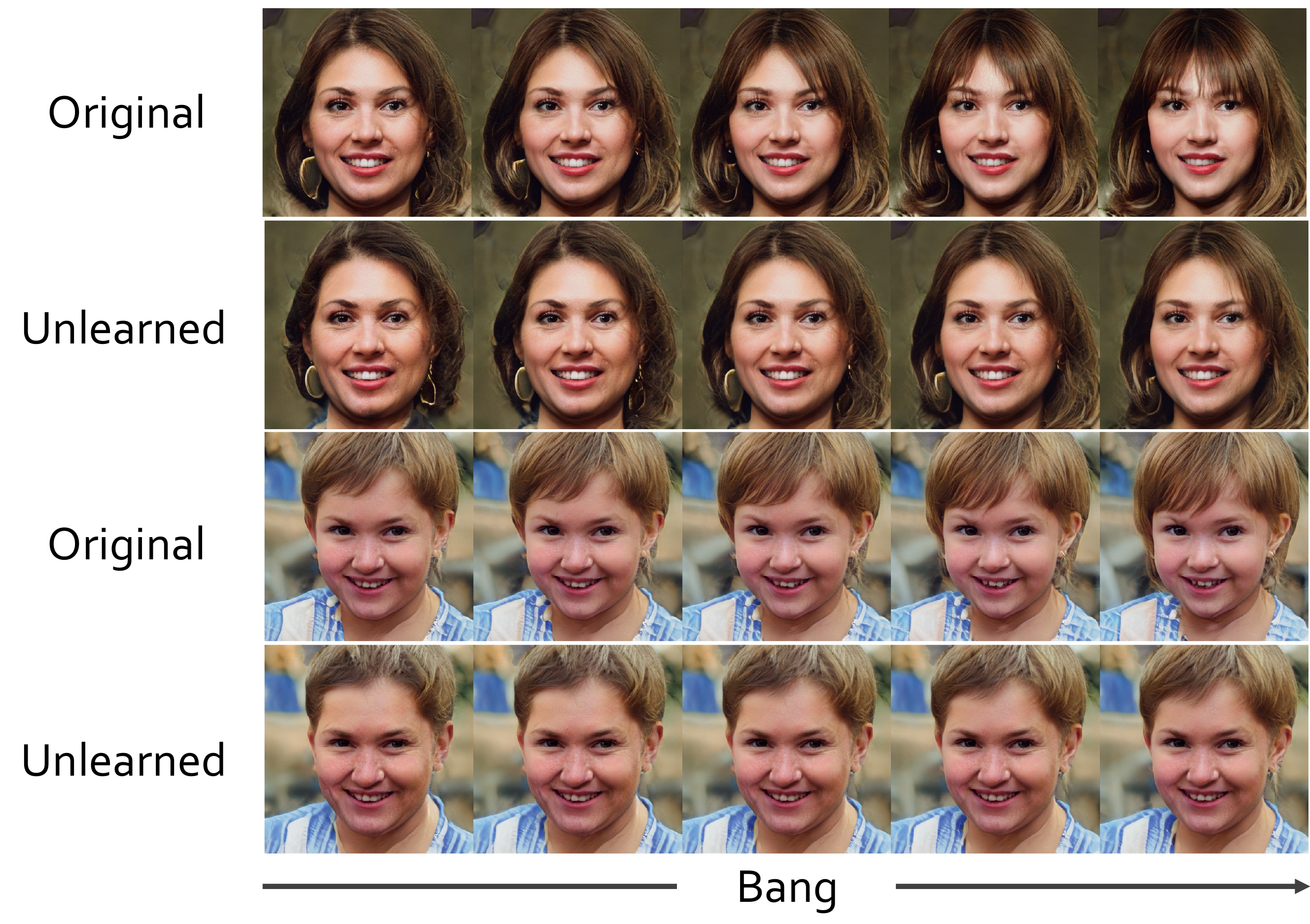}
    \caption{Visualization of `Bang' feature before and after unlearning from pre-trained StyleGAN model. All paired images in each column are generated from the same latent vector.}
\end{figure}
\begin{figure}[h!]
    \centering
    \includegraphics[width=0.76\linewidth]{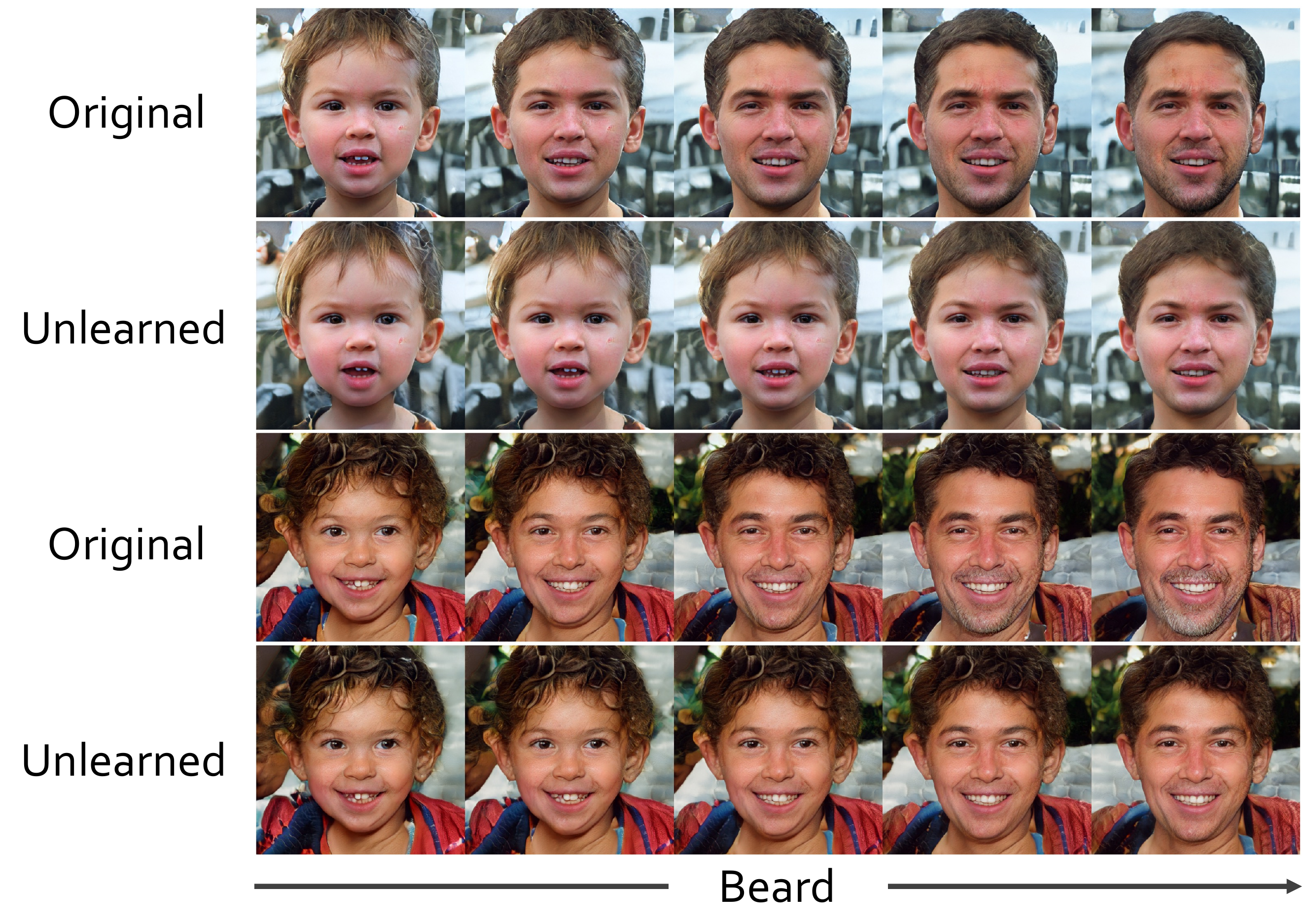}
    \caption{Visualization of `Beard' feature before and after unlearning from pre-trained StyleGAN model. All paired images in each column are generated from the same latent vector.}
\end{figure}
\begin{figure}[h!]
    \centering
    \includegraphics[width=0.76\linewidth]{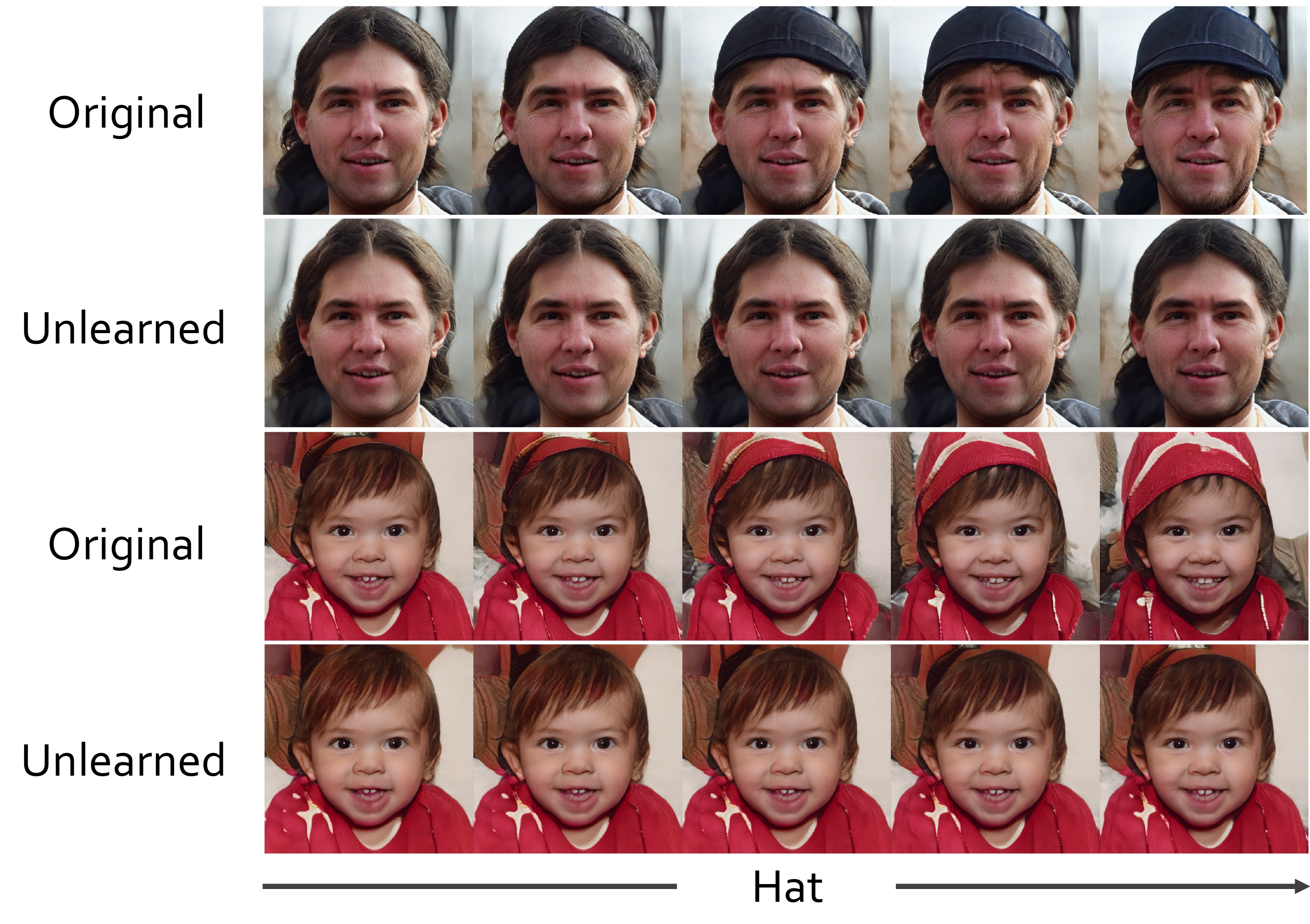}
    \caption{Visualization of `Hat' feature before and after unlearning from pre-trained StyleGAN model. All paired images in each column are generated from the same latent vector.}
\end{figure}
\begin{figure}[h!]
    \centering
    \includegraphics[width=0.76\linewidth]{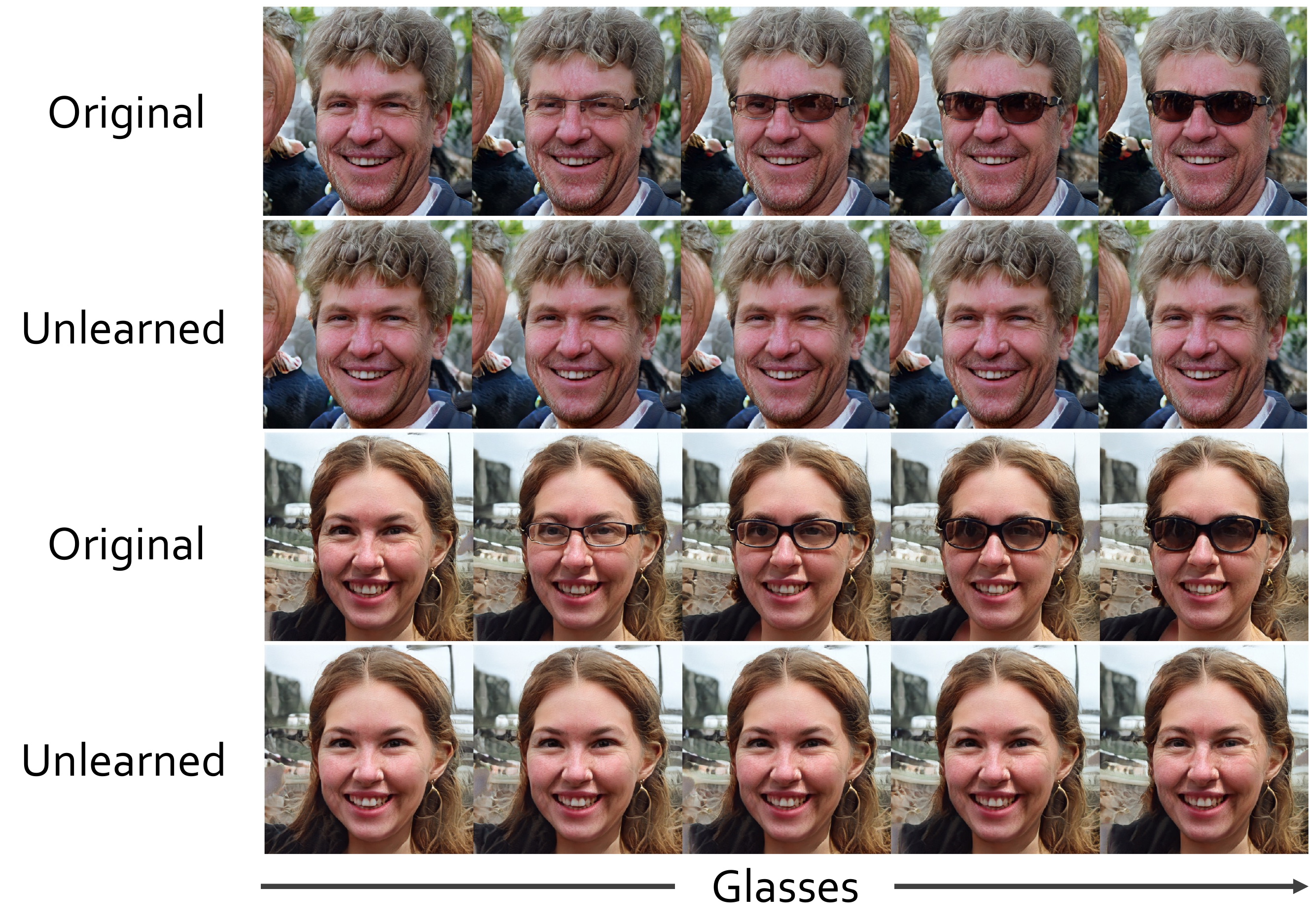}
    \caption{Visualization of `Glasses' feature before and after unlearning from pre-trained StyleGAN model. All paired images in each column are generated from the same latent vector.}
\end{figure}

\subsection{Analysis of the experiment}
\paragraph{Target identification in latent space.}
The feature identification method proposed in Eq. \ref{eqn:feature_cls} raises a question about the quality of the result. We use the same classifier used to measure the target feature ratio to measure the quality of the feature identification method. Note that the pre-trained classifiers are only used in the evaluation and not given during unlearning.

Table \ref{table:similarity} shows the ROC-AUC score of the feature identification used in each experiment. The feature classification method achieves relatively high accuracy without an external classification model showing that the feature extracted from the latent space can be used for unlearning. 
\begin{table}[h!]
\centering
\begin{tabular}{ccc}
\toprule
\multicolumn{3}{c}{MNIST}  \\ \midrule
Models & Thickness & Slant \\ \cmidrule{1-3}
VAE    &  0.908  &   0.936    \\
DCGAN  &   0.853        &   0.858    \\ \midrule
\multicolumn{3}{c}{CelebA} \\ \midrule
Models & Bang     & Beard \\ \cmidrule{1-3}
VDVAE  &0.840&  0.834     \\ 
ProgGAN  &  0.891      &  0.806     \\ \bottomrule
\end{tabular}
\caption{ROC-AUC score of feature identification method.}
\label{table:similarity}

\end{table}


\paragraph{Ablation on the objective.}
We conduct an ablation study to evaluate the effectiveness of our proposed objective function. Specifically, we experiment with erasing `Bang' in a pre-trained GAN trained with the CelebA dataset. Table \ref{table/ablation} provides a detailed comparison of the performance under different combinations of objectives. Without the perception loss, the model achieves a better unlearning performance in terms of target ratio, sacrificing the quality of images.

\begin{table}[h!]

\centering
\begin{tabular}{cccrrr}
\toprule
$\mathcal{L}_{\text{recon}}$ & $\mathcal{L}_{\text{unlearn}}$ & $\mathcal{L}_{\text{percep}}$ & TFR (\%) & IS & FID\\ \midrule
      & $\checkmark$       &        &   0.08   & 2.89 &52.48\\
$\checkmark$     & $\checkmark$       &        &    0.38   & 2.85 &50.93\\
$\checkmark$     & $\checkmark$       & $\checkmark$      &    0.42 & 2.92  & 49.82   \\\midrule
\multicolumn{3}{c}{Baseline} &      0.49       &2.91 &    51.37\\\midrule
\multicolumn{3}{c}{Original} &  6.74  &2.92& 48.05  
\\
\bottomrule
\end{tabular}
\caption{Ablation study to unlearn the `Bang' feature from a pre-trained GAN. Each metric indicates the target feature ratio ($\downarrow$), inception score ($\uparrow$), and Fréchet inception distance ($\downarrow$).}
\label{table/ablation}
\end{table}

\paragraph{Choices of hyper-parameter.}
We analyze the results with varying hyper-parameter $\alpha$ on unlearning the `Bang' feature from GAN. As shown in Table \ref{table/alpha}, the importance of the unlearning objective becomes more significant as $\alpha$ increases. Consequently, increasing $\alpha$ results in a decrease in the target ratio, which successfully erases the target feature from the generated images. However, we observe the trade-off between the target feature ratio and the quality of the generated images. Therefore, careful selection of $\alpha$ is important to achieve the desired balance between effective unlearning and preserving image quality. 
\begin{table}[h!]
\centering
\begin{tabular}{crrr}
\toprule
 & Target feature ratio ($\downarrow$) & IS ($\uparrow$)    & FID ($\downarrow$)    \\ \midrule
$\alpha$ = 1 & 5.36                & 2.87 & 48.41 \\
$\alpha$ = 2 & 1.08                & 2.89 & 49.09 \\
$\alpha$ = 3 & 0.42                & 2.91 & 49.82 \\
$\alpha$ = 4 & 0.21                & 2.89 & 49.93 \\
$\alpha$ = 5 & 0.01                & 2.87 & 50.57 \\ \bottomrule
\end{tabular}
\caption{Result of hyper-parameter analysis varying $\alpha$}
\label{table/alpha}
\end{table}


\section{User Study}
\label{sec:user_detail}
\subsection{User Interface.}
To conduct the user study, we recruit 13 participants from a graduate school studying machine learning and artificial intelligence.
Participants first identify generated images with specific target features (e.g., Glasses). Using the collected dataset, we unlearn pre-trained StyleGAN with our proposed unlearning framework. Then, each participant assesses their customized unlearned model based on three criteria: target feature ratio, image quality, and pinpoint unlearning. This section shows the screenshot from the user study discussed in Section~\ref{sec:user_study}.

\begin{figure}[h!]
    \centering
    \includegraphics[width=0.43\linewidth]{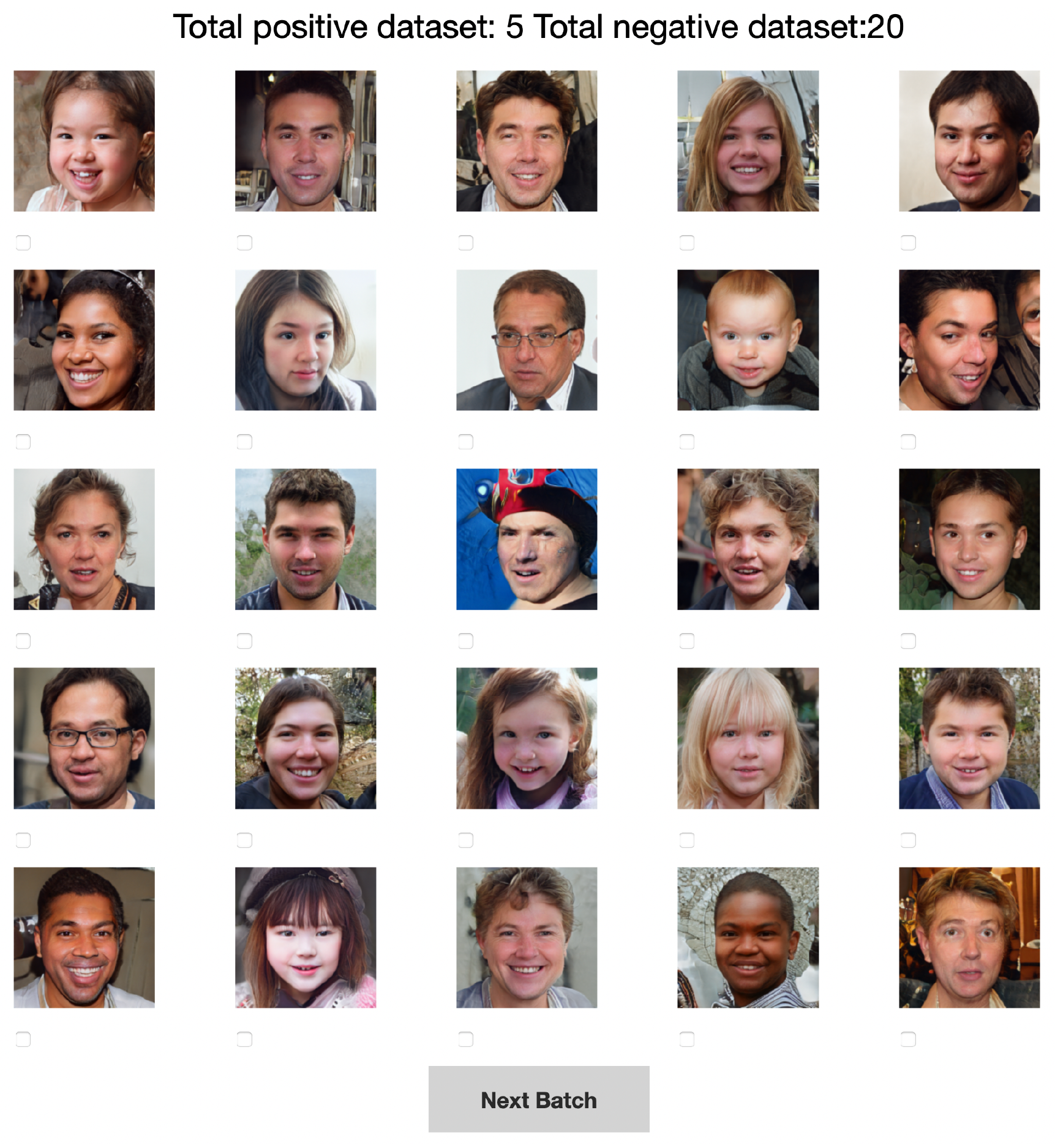}
    \caption{Screenshot for annotating images that have target feature. Participants examine 500 images from generated images.}
\end{figure}
\begin{figure}[h!]
    \centering
    \includegraphics[width=0.43\linewidth]{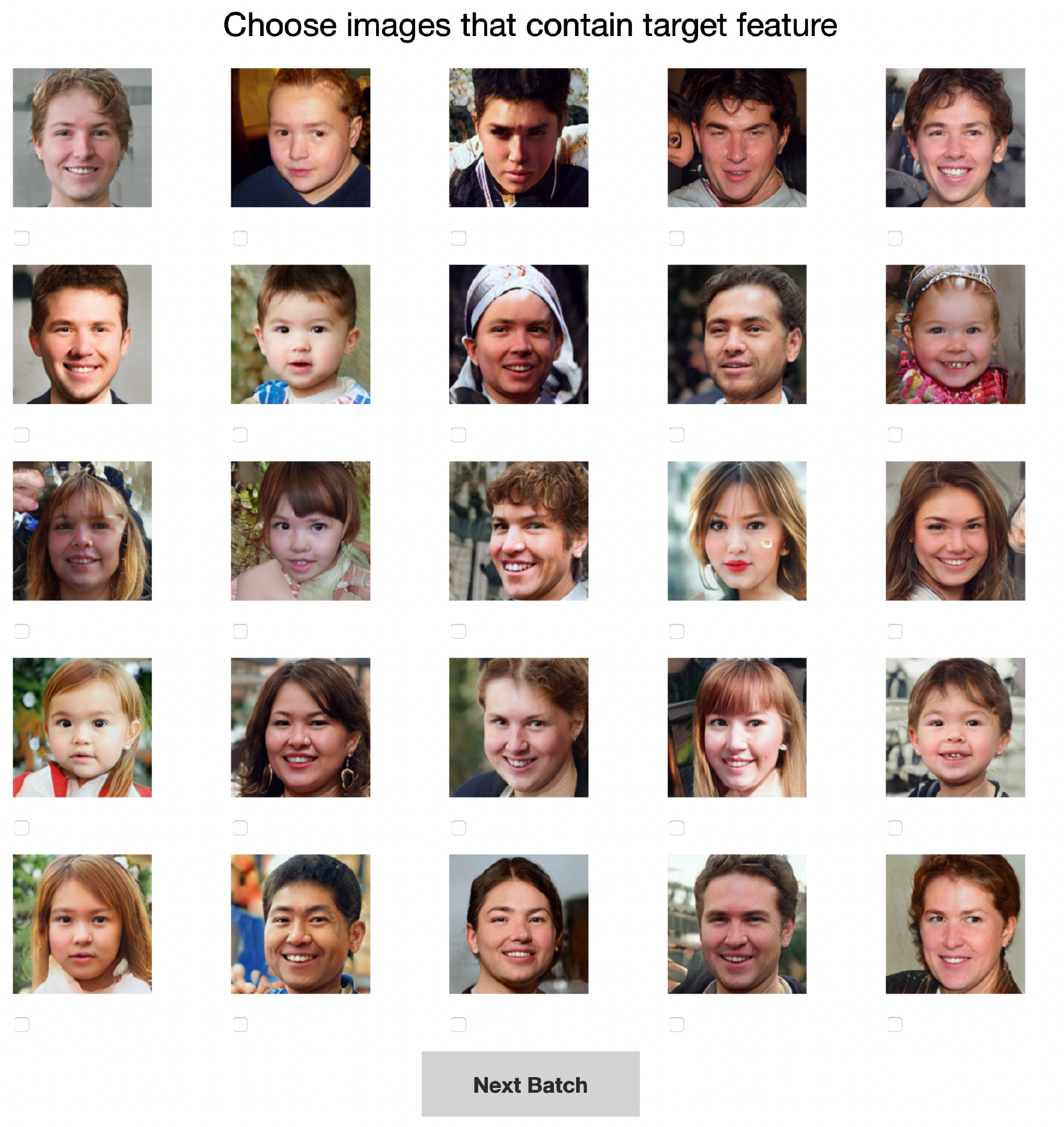}
    \caption{Screenshot for evaluating target feature ratio. Given 500 images, participants select the images that have a target feature.}
\end{figure}
\begin{figure}[h!]
    \centering
    \includegraphics[width=0.5\linewidth]{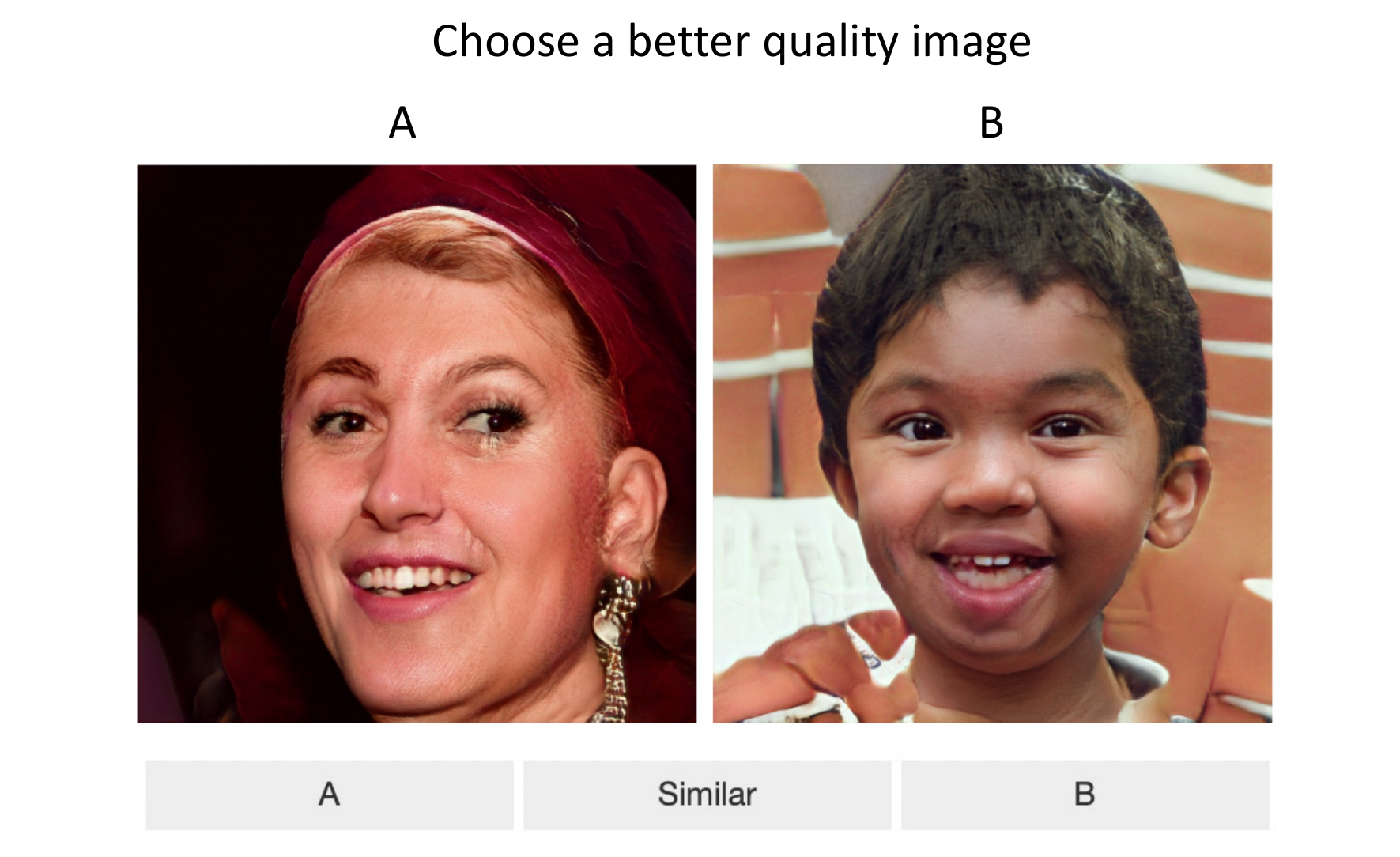}
    \caption{Screenshot for evaluating image quality. Participants choose the better quality image given two random images generated from the original model and the unlearned model.}
\end{figure}
\begin{figure}[h!]
    \centering
    \includegraphics[width=0.5\linewidth]{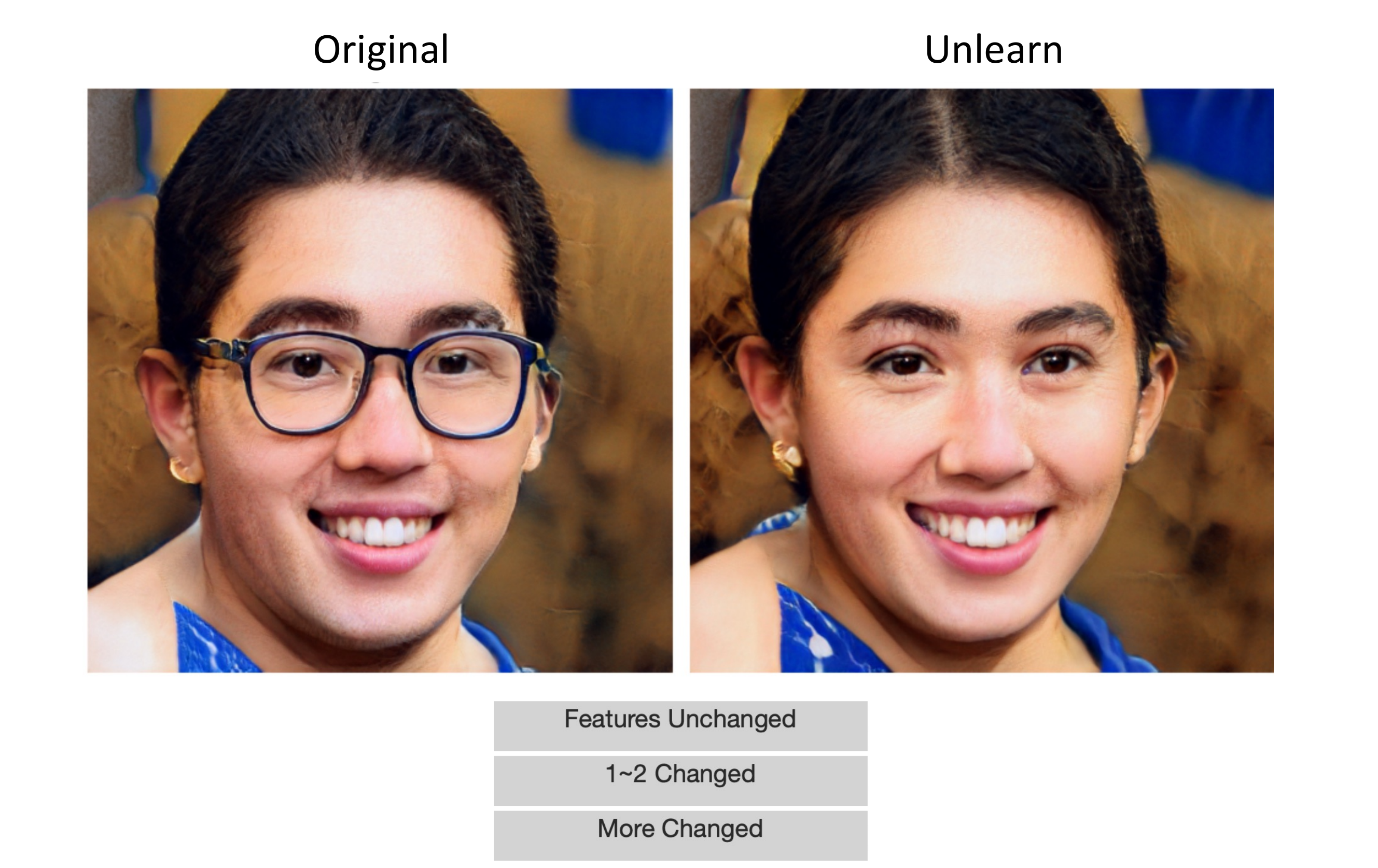}
    \caption{Screenshot for evaluating pinpoint unlearning. Participants are provided with two images: one from the original model and another from the unlearned model. Both are generated using the same latent vector. Then they measure how many features are changed except the target feature.}
\end{figure}

\subsection{Detail results of user study}
Table~\ref{tab:userstudy} provides the average and standard deviation of three user study experiments.
\begin{table}[h!]
\centering
\begin{tabular}{cc r}
\toprule
Experiments & Source & Avg.$\pm$Std.\\ \midrule
\multirow{2}{*}{Target feature ratio} & Original         & 47.92$\pm$6.04 \\
                                      & Unlearn          & 3.23$\pm$1.92  \\ \midrule
\multirow{3}{*}{Image quality}        & Original         & 15.15$\pm$5.10 \\
                                      & Unlearn          & 15.23	$\pm$4.54 \\
                                      & Similar          & 19.61$\pm$7.46 \\ \midrule
\multirow{3}{*}{Pinpoint unlearning}  & Unchanged        & 4.53$\pm$2.26  \\
                                      & 1$\sim$2 changed & 4.38$\pm$1.66  \\
                                      & More changed     & 1.07$\pm$1.80  \\ \bottomrule
\end{tabular}
\caption{Average and standard deviation of user study result.}
\label{tab:userstudy}
\end{table}

\section{PGD Attack}
\label{sec:PGD_attack}
We employ a Projected Gradient Descent (PGD) attack~\cite{madry2017towards} on the latent variable of the unlearned model in Section~\ref{sec:adversarial_attack}. The attack is guided by a pre-trained feature classifier capable of classifying the target feature. The purpose of this attack is to determine whether the unlearned model can be manipulated to produce images that contain the target feature, even after it has been supposedly removed. 
Specifically, let $h:\mathcal{X}\rightarrow [0,1]$ be the target feature classifier. With randomly sampled latent vector $\rvz$, the PGD attack aims to find $\tilde\rvz$ such that
\begin{align*}
    \tilde\rvz = \argmin_{\rvz' \in \Delta_\rvz}\operatorname{CE}(1, h(g_\theta(\rvz'))),
\end{align*}
where $\operatorname{CE}$ is a cross-entropy loss, and $\Delta_\rvz$ defines a set of possible perturbations from the original value $\rvz$. Hence, the PGD attack finds the latent variable generating a sample that can be classified as the one with the target feature. The overall method used in this experiment is provided in Algorithm \ref{alg:attack}.
\begin{algorithm}[h!]
    \caption{Projected Gradient Descent method}
    \label{alg:attack}
    \begin{algorithmic}
        \REQUIRE $h$: target classifier, $g_\theta$: unlearned model
        \REQUIRE $A$: attack step, $\eta$: learning rate, $\epsilon$: maximum magnitude of perturbation
        \STATE Sample a latent vector $\rvz$ from $N(0,1)$
        \STATE Initialize a perturbation $\rvz' = \rvz$ 
        \FOR{$i=1$ {\bfseries to} $A$}
            \STATE Compute the loss $\mathcal{L} = \operatorname{CE}(1, h(g_\theta(\rvz')))$
            \STATE Compute gradient $\nabla = \frac{\partial \mathcal{L}}{\partial \rvz'}$
            \STATE Update the perturbation $\rvz' = \rvz' + \eta \cdot \operatorname{sign}(\nabla)$
            \STATE Project $\rvz'$ to $\epsilon$-ball under the infinity norm: 
            \STATE $\rvz' = \min(\max(\rvz'-\rvz, -\epsilon), \epsilon)$
        \ENDFOR
        \RETURN $\rvz'$
    \end{algorithmic}
\end{algorithm}

\end{document}